\documentclass[runningheads]{llncs}

\usepackage{eccv}

\usepackage{eccvabbrv}

\usepackage{graphicx}
\usepackage{booktabs}
\usepackage{amsmath}
\usepackage{amssymb}
\usepackage{algorithm}
\usepackage{algpseudocode}   
\usepackage{placeins}

\usepackage[accsupp]{axessibility}  

\usepackage[pagebackref]{hyperref}

\usepackage{orcidlink}

\begin{document}

\title{Mind-to-Face: Neural-Driven Photorealistic Avatar Synthesis via EEG Decoding\thanks{Project page: \url{https://tianwen-fu.github.io/mind_to_face}} } 
\titlerunning{Mind-to-Face}
\authorrunning{H.~Xiong et al.}

\author{
Haolin Xiong$^{*}$ \and
Tianwen Fu$^{*}$ \and
Pratusha Bhuvana Prasad \and
Yunxuan Cai \and
Wenbin Teng \and
Haiwei Chen \and
Hanyuan Xiao \and
Yajie Zhao$^{\dagger}$
}
\institute{University of Southern California \\
USC Institute for Creative Technologies\\
\textsuperscript{*}Equal contribution.\quad
\textsuperscript{$\dagger$}Corresponding author.}

\maketitle
\begin{center}
    \centering
    \captionsetup{type=figure}
    \includegraphics[width=0.95\textwidth]{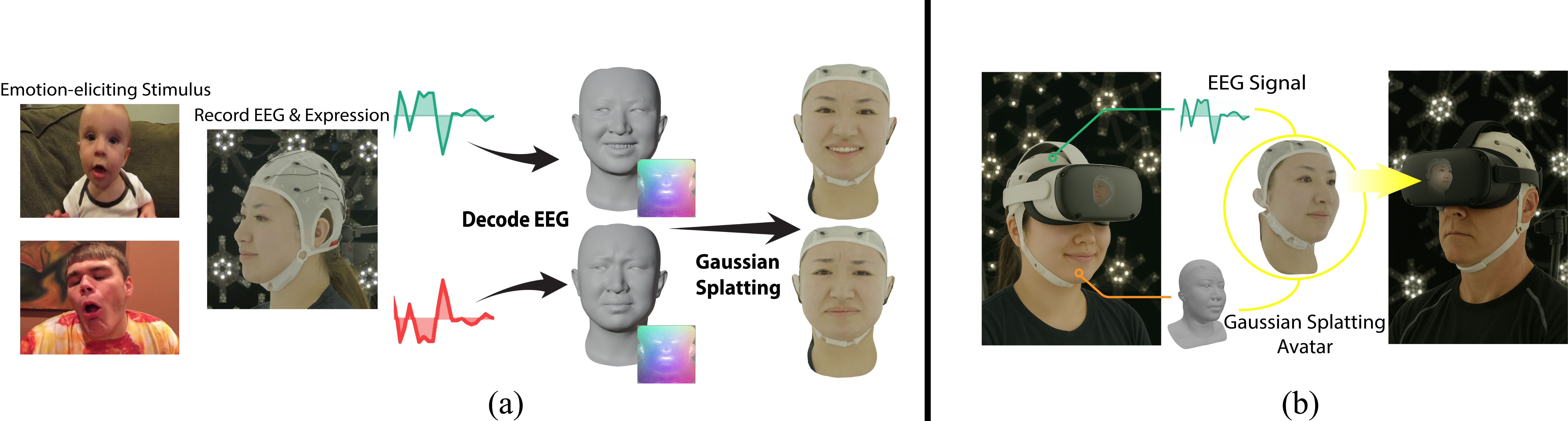}
    \captionof{figure}{Overview of Mind-to-Face, our neural-driven avatar framework. (a) We record synchronized EEG and facial expressions while subjects view emotion-eliciting stimuli, then decode the EEG into dense 3D position maps, and render photorealistic avatars using 3D Gaussian Splatting~\cite{kerbl20233dgaussiansplattingrealtime}.
    (b) While exploratory and not intended as a complete VR system, this work suggests an application: decoding EEG to drive expressive avatars when the face is occluded by a head-mounted display. Mind-to-Face points toward future systems that enable expressive avatar control under full facial occlusion.}
    \label{fig:teaser}
\end{center}

\begin{abstract}
Current expressive avatar systems rely heavily on visual cues and often fail when faces are occluded or emotions remain internal. We present \textbf{Mind-to-Face}, the first framework to decode non-invasive electroencephalogram (EEG) signals directly into high-fidelity facial expressions. We build a dual-modality recording setup that captures synchronized EEG and multi-view facial video during emotion-eliciting stimuli, providing precise supervision for neural-to-visual learning. Our model uses a CNN-Transformer encoder to map EEG signals into dense 3D position maps that sample over 65k vertices, capturing fine-scale geometry and subtle emotional dynamics, and renders them through a modified 3D Gaussian Splatting pipeline for photorealistic, view-consistent results. Extensive evaluations show that EEG alone can reliably predict dynamic, subject-specific facial expressions, including subtle emotional responses, demonstrating that neural signals contain far richer affective and geometric information than previously assumed. \textbf{Mind-to-Face} establishes a new paradigm for neural-driven avatars, enabling personalized, emotion-aware telepresence and cognitive interaction in immersive environments.
\keywords{Neural Decoding \and Affective Computing \and EEG \and BCI \and Gaussian Avatar \and 3DGS \and Light Stage}
\end{abstract}

\section{INTRODUCTION}
\label{sec:introduction}
Current avatar and emotion systems rely almost entirely on visual cues, such as cameras, motion tracking, or speech analysis, to infer and render human expressions. These approaches fail when faces are occluded by Head-mounted Displays (HMDs) or when emotions are internally experienced but not outwardly expressed. This limitation overlooks a fundamental fact: human emotion and cognition originate from neural activity that precedes facial movement and behavior. Yet current research treats these modalities separately—vision models reconstruct external geometry, affective models classify EEG into coarse labels (e.g., ``happy'' or ``sad''), and brain-computer interfaces (BCIs) focus on symbolic or motor signals. A critical gap remains in linking neural states to visual expressions. Bridging this gap would enable emotionally transparent communication when visual channels are blocked, EEG-driven avatars for gaming and collaboration, adaptive avatars that reflect personalized neural patterns, and new neuro-visual datasets for decoding attention, emotion, and memory into visual form.

Electroencephalography (EEG) is a non-invasive method for measuring electrical brain activity and is widely used in brain–computer interfaces (BCIs). Prior work demonstrates strong correlations between EEG signals and emotional states~\cite{Duan2013, Zheng2018} or symbolic facial expressions~\cite{Zheng2017}, indicating their potential for decoding expressive information directly from brain activity. However, despite increasing interest in EEG-based applications, EEG has rarely been used as a driver for visual synthesis. Recent learning-based approaches classify discrete expression labels from EEG~\cite{Li2020, Li2022}, but remain limited to coarse categories and cannot capture the continuous, fine-grained dynamics required for realistic avatar control. Achieving high-fidelity reconstruction from EEG is challenging due to the low dimensionality and noisiness of neural signals.

We present Mind-to-Face, a framework that bridges neural activity and high-fidelity facial expression, as an exploratory method to enable expression decoding from brain activity rather than what cameras capture. Our approach addresses three core challenges. First, we design a dual-modality data acquisition setup that synchronously records EEG signals and multi-view facial video during emotion-eliciting stimuli, solving a key engineering challenge in aligning neural activity with visual expression. Second, we introduce a neural decoding pipeline that translates EEG patterns into photorealistic renderings via 3D Gaussian Splatting. 

Unlike images used in traditional face tracking, neural responses are highly individualized: identical stimuli can produce distinct EEG signatures across subjects. Therefore, our work focuses on subject-specific expression control rather than scalable cross-subject modeling. In this sense, our approach follows the spirit of prior BCI research (e.g.,~\cite{Willett2023.01.21.524489}), where evaluation on a small number of participants is common at exploratory stages. Accordingly, we collect a dual-modality dataset from two representative subjects across six trials, each containing multiple stimulus videos. Evaluating two subjects allows a controlled study of neural-to-geometry mapping quality, providing a foundation for future work on cross-subject generalization.

To extract features from EEG responses, we adopt a CNN-Transformer hybrid encoder inspired by EEG-Conformer~\cite{song2023eeg}, capturing both local spatial correlations and long-range temporal dependencies in EEG. Each EEG slice is encoded into a latent representation of brain activity, which is then decoded into a dense 3D position map~\cite{liu2022rapid} to produce continuous facial geometry in UV space. Unlike strain-gauge systems~\cite{li2015}, EEG offers a richer and more holistic measure of cognitive and affective states. For photorealistic rendering, we build on the 3D Gaussian Splatting framework~\cite{kerbl20233dgaussiansplattingrealtime} following GaussianAvatars~\cite{qian2024gaussianavatarsphotorealisticheadavatars}, but replace the low-dimensional FLAME model~\cite{FLAME2017} with a high-degree-of-freedom 3D position map. This dense representation captures subtle geometric details and fine expression nuances that are smoothed out in low-dimensional models. We further modify the deformation and skinning system of GaussianAvatars to support this representation and incorporate a robust facial-tracking pipeline during preprocessing, enabling accurate reconstruction even under occlusion and challenging expressions. 

In summary, our key contributions include: 
\begin{enumerate}
\item We introduce \textit{Mind-to-Face}, the first system that decodes non-invasive EEG signals into dense 3D facial geometry and renders photorealistic avatars. Our approach establishes a continuous neural-to-visual mapping, enabling expressive facial motion to be reconstructed directly from brain activity.

\item We design a custom \textbf{dual-modality data acquisition system} that synchronously records 16-channel EEG signals and high-speed multi-view facial videos while subjects watch emotion-eliciting film clips. This setup provides precise temporal alignment between neural activity and facial responses, enabling supervised learning for neural-driven 3D facial synthesis. We will publicly release the resulting synchronized EEG-to-3D dataset, which to the best of our knowledge is the first of its kind, supporting future research in neural decoding, telepresence, and affective computing.

\end{enumerate}

\section{RELATED WORK}
\label{sec:related}
\subsection{EEG Decoding and Emotion Recognition}
Electroencephalography (EEG) decoding is an active area of research in brain-computer interface (BCI), emotion recognition, and cognitive state estimation~\cite{sun2023survey}. Traditional approaches rely on handcrafted features such as band power~\cite{ZHANG2023104157}, Common Spatial Patterns (CSP)~\cite{10405666}, or functional connectivity~\cite{WOLPAW202015} to summarize EEG into low-dimensional descriptors, often limiting the richness of information available for learning expressive behaviors. Recent advances leverage deep learning to directly extract spatial and temporal features from raw EEG. Convolutional networks (CNNs) are widely used for spatial filtering across channels~\cite{Lawhern_2018,RIYAD2021109037,song2023eeg}, while RNNs~\cite{postepski2024recurrentconvolutionalneuralnetworks} and Transformers~\cite{song2023eeg,song2021transformerbased} have been explored for temporal modeling. 

There are two primary theoretical models for representing human emotions. The first is the discrete model, which categorizes emotions into a fixed set of basic classes-originally six, as introduced by Ekman~\cite{ekman1992argument}, and later expanded to include up to fifteen~\cite{Nadeem24}. In contrast, the dimensional model represents emotions along continuous axes such as arousal, valence, pleasure, displeasure, sleepiness, and distress, as proposed by Russell~\cite{russell1980circumplex}. These models have been widely adopted in emotion recognition research and dataset design.

Early EEG-based emotion recognition systems~\cite{Liu2010, Sollfrank2021} visualized categorical emotional states using digital avatars. The SEED dataset~\cite{zheng2015SEED} introduced a collection of EEG recordings for tasks including sleep state detection, eye blink detection, and emotion classification. The publicly available dataset most similar to our setting is MAHNOB-HCI~\cite{MAHNOB}, which includes one RGB and five monochrome facial video streams synchronized with 32-channel EEG recordings. However, due to its low resolution and monochrome format, MAHNOB-HCI cannot be used for 3D facial reconstruction. To the best of our knowledge, our Mind-to-Face data capture setting is the first to collect a high-fidelity multi-view dataset that enables photogrammetric 3D face reconstruction synchronized with EEG signals as well as stimuli videos.

\subsection{Head Reconstruction and Neural Avatars}
Recent advances in differentiable rendering and implicit scene representations have substantially improved the realism and controllability of digital head avatars. Thies et al.~\cite{Thies2016} introduced a real-time face tracking and transfer system, establishing the foundation for neural avatar animation. Subsequent progress in face image synthesis~\cite{Chan2018, Thies2019} has enabled more expressive control over facial expressions, lip synchronization, and head motion~\cite{Suwajanakorn2017, kim2018deep, Xu2023}.

More recently, Neural Radiance Field (NeRF)–like methods~\cite{mildenhall2020nerfrepresentingscenesneural} have since emerged as a powerful paradigm for realistic head avatar modeling and rendering. Gafni et al.~\cite{Gafni_2021_CVPR} learn a dynamic NeRF conditioned on expression vectors from monocular video input, while Grassal et al.~\cite{grassal2022neuralheadavatarsmonocular} extend the FLAME model~\cite{FLAME2017} by subdividing the mesh and introducing expression-dependent geometric and texture offsets to enhance realism. Other approaches, such as IMavatar~\cite{zheng2022imavatar}, employ neural implicit functions to model avatars and solve for deformations between observed and canonical poses via iterative root-finding.

Since 2023, 3D Gaussian Splatting (3DGS)~\cite{kerbl20233dgaussiansplattingrealtime} has emerged as a powerful yet efficient alternative to NeRF-like methods. A growing body of work has explored 3DGS for human avatar animation, including controllable full-body avatars~\cite{hu2024gaussianavatar, zielonka2025drivable3dgaussianavatars}, motion synthesis~\cite{moon2024expressivewholebody3dgaussian, yuan2023gavatar}, and head reconstruction~\cite{tang2025gafgaussianavatarreconstruction}. Notably, GaussianAvatars~\cite{qian2024gaussianavatarsphotorealisticheadavatars} introduced a framework that combines FLAME-based explicit expression control with 3DGS rendering, achieving high-fidelity geometry and photorealistic appearance through anisotropic splatting. Importantly, while preserving the fine-grained visual fidelity of NeRFs, 3DGS offers an explicit scene representation that enables precise control over geometry and appearance, making it particularly well-suited for animation tasks.

Our work builds upon the rendering paradigm of GaussianAvatars~\cite{qian2024gaussianavatarsphotorealisticheadavatars}. Replacing the traditional blendshape and FLAME-based geometric control with a dense, per-pixel 3D position map representation, our method captures fine-grained facial details and subtle expression dynamics with significantly higher fidelity.

\section{METHOD}
\begin{figure*}[ht]
    \centering
    \includegraphics[width=\linewidth]{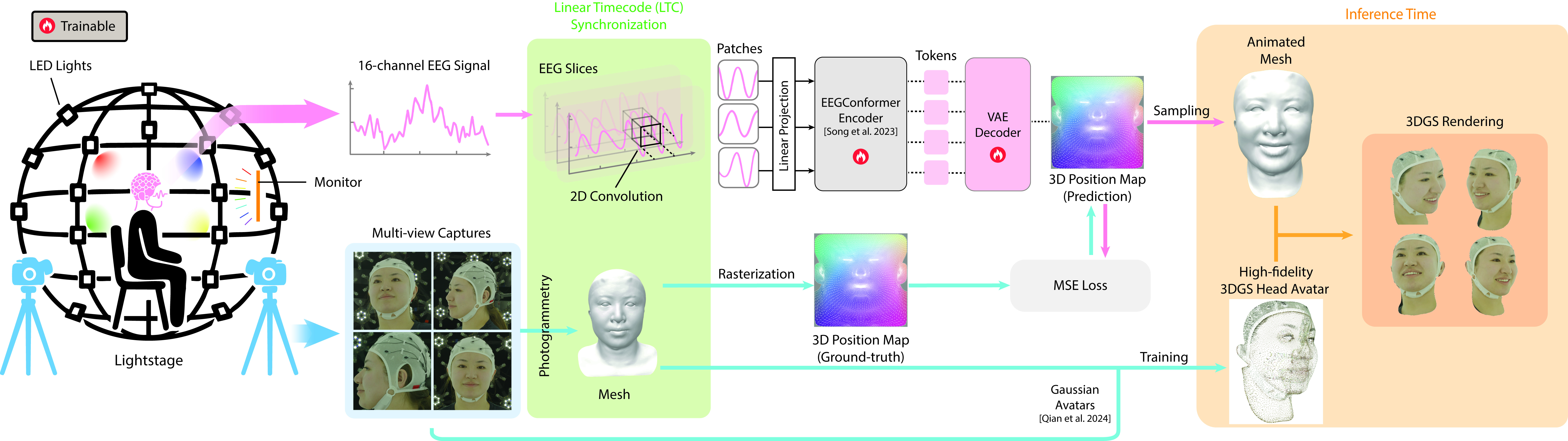}
    \caption{\textbf{Overview of Our Mind-to-Face Pipeline.~}Our system decodes raw EEG signals into dense 3D position maps that are rendered into photorealistic avatars. Data are collected using a custom multi-view capture rig composed of synchronized high-speed RGB cameras and a 16-channel EEG headset, with all modalities temporally aligned via Linear Timecode for frame-accurate correspondence between EEG and facial expressions. Multi-view videos are reconstructed into ground-truth 3D facial meshes through photogrammetry. During training, EEG slices are encoded using a CNN–Transformer encoder (EEG-Conformer~\cite{song2023eeg}) and decoded into 3D position maps using a randomly initialized VAE decoder, supervised with MSE loss against photogrammetric position maps. At inference, the predicted position maps are resampled into meshes and rendered using a modified GaussianAvatars pipeline~\cite{qian2024gaussianavatarsphotorealisticheadavatars} for high-fidelity avatar synthesis.}
    \label{fig:pipeline}
\end{figure*}
\label{sec:method}
Our objective is to decode continuous facial expressions from EEG signals. To achieve this, we first design a multi-modal data capture system (\cref{sec:DataCaptureSetup}) and perform synchronized EEG-video data acquisition (\cref{sec:Acquisition}) followed by signal and geometry preprocessing (\cref{sec:preprocessing}). Using the processed data, we develop an EEG encoder framework based on a convolutional transformer architecture~\cite{song2023eeg}, coupled with an decoder~\cite{rombach2021highresolution} that learns to map brain signals to facial geometry in the format of positional map(\cref{sec:conformer}). Finally, the reconstructed EEG-driven meshes are rendered photorealistically using a variant of GaussianAvatars pipeline~\cite{qian2024gaussianavatarsphotorealisticheadavatars}, enabling realistic and view-consistent avatar synthesis. Our overall framework is illustrated in~\cref{fig:pipeline}.

\subsection{Data Capture Setup}
\label{sec:DataCaptureSetup}

\paragraph{\bf{Facial Performance Capturing.}} 
We record a participant’s responses to stimulus videos from multiple views under a constant lighting contition. Sixteen cinema-grade global-shutter cameras (up to 8K resolution, 120 fps) are evenly distributed around the subject at 15\textdegree~intervals, providing dense view coverage for photogrammetric reconstruction. A display positioned in front of the participant presents emotionally evocative video stimuli to elicit natural expressions. An illustration of our capture setup is shown in the supplementary materials. All cameras are synchronized using genlock and timecode generators to ensure precise temporal alignment across all views for high-fidelity reconstruction.

\paragraph{\bf{EEG Capturing Device.}}
We use the Cyton-Daisy biosensing kit by OpenBCI for EEG data acquisition due to its portability and reliable signal quality. The device records 16-channel EEG signals at a sampling rate of 125Hz. Electrodes are positioned according to the international 10–20 system, as illustrated in the supplementary materials, providing evenly distributed coverage across the scalp. Although the number of channels is fewer than that of clinical-grade EEG systems, this configuration offers sufficient spatial resolution to capture representative neural activity across major cortical regions~\cite{Ameri2018, Jiang2019}, making it well-suited for our expression decoding task. Conductive gel is applied to each electrode to maintain impedance below 500k$\Omega$, ensuring stable and high-quality signal acquisition.

\paragraph{\bf{Synchronization.}}
Synchronization between the camera array and the EEG headset is achieved through a linear signal generated by an Arduino Teensy board, ensuring frame-accurate temporal alignment between EEG measurements and corresponding facial expressions. This setup yields a dual-modality dataset with precisely synchronized EEG and multi-view facial recordings, enabling reliable cross-modal learning.

\subsection{Experiments Design and Data Acquisition}
\label{sec:Acquisition}
\paragraph{\bf{Stimulus Videos.}} The stimulus videos are manually selected from the Open Library for Affective Videos (OpenLAV)~\cite{Israel2021} and emotion eliciting clips curated by Coan and Allen~\cite{coan2007handbook}. The selected clips are grouped into five emotion specific categories—neutral, disgust, funny, angry, and sad—each constituting a single trial in subsequent experiments. In addition, a separate video containing mixed emotional content is compiled from the EMOSTIM dataset~\cite{emostim} and used as a leave-one-trial-out testing sequence.

\begin{figure}[htbp]
    \centering
    \includegraphics[width=0.8\linewidth]{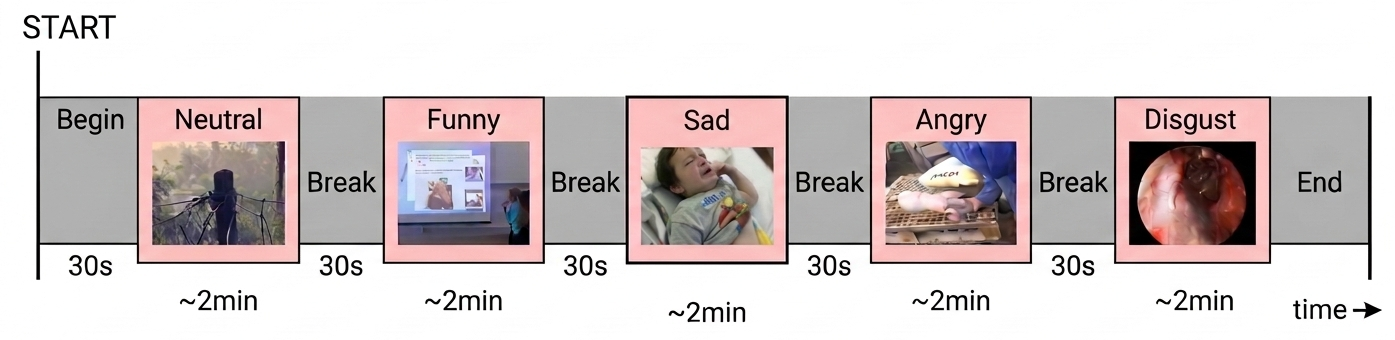}
    \setlength{\belowcaptionskip}{-10pt}
    \caption{\textbf{Trial Design.}~The experiment consists of five emotion-specific trials, each separated by a 30-second rest period. Within each trial, participants view a sequence of video clips conveying a single, consistent emotion corresponding to that trial's theme.}
    \label{fig:trials}
\end{figure}

\paragraph{\bf{Experiment Procedure.} }
The subject is seated at the center of the Light Stage while wearing the EEG headset. The stage is uniformly illuminated with a high depth-of-field lighting configuration to minimize motion blur during performance capture. An illustrative example of the capture setup is shown in the supplementary materials. Stimulus playback, EEG recording, and multi-view video capture are initiated simultaneously for precise temporal alignment. At the beginning of each trial, a prompt appears on the display indicating the target emotion category, preparing the subject’s mental state for the upcoming stimuli. The corresponding stimulus videos are then presented, followed by a 30-second rest period between trials to prevent emotional carryover effects. Each subject completes five stimulus trials, as illustrated in \cref{fig:trials}, resulting in approximately 25 minutes of total recording time. The setup yields a synchronized dual-modality dataset comprising high-resolution multi-view videos and EEG signals. In total, around 30 minutes video data per-subject is captured in compressed RAW format from 16 synchronized cameras, producing roughly 7TB of data per subject. Despite the dataset’s large size and preprocessing demands, all recordings were successfully processed and prepared for decoding.

\subsection{Preprocessing}
\label{sec:preprocessing}
\paragraph{\bf{EEG Preprocessing.}}\label{par:EEG_preprocessing} For each trial, the raw EEG data is represented as a time-varying signal with $C$ channels sampled at each time step. A 6\textsuperscript{th}-order Butterworth band-pass filter with a passband of 4-40Hz is applied to suppress high-frequency noise and remove low-frequency drift and DC components introduced by the recording hardware. Per-channel z-score standardization is then performed using statistics computed from the training set, with the corresponding mean and variance applied to the test set for normalization consistency. The filtered and normalized EEG signals are subsequently segmented into overlapping temporal windows, each denoted as $E^{(t)} \in \mathbb{R}^{W \times C}$, where $W=375$ is the window size, and $C=16$. Each EEG window $E^{(t)}$ is temporally aligned with a corresponding ground-truth 3D position map $P^{(t)}$ reconstructed via multi-view photogrammetry.

\paragraph{\bf{Face Geometry Reconstruction.} }
To enable supervised learning for decoding facial expressions, we reconstruct a 3D facial mesh $M^{(t)}$ for each video frame $I^{(t)}$ captured during the trial sessions. We track the subject's performance by first solving for the facial identity of a neutral expression frame using the 3D Morphable Model (3DMM) from ICT-FaceModel~\cite{li2020learning}, and then fitting the identity template mesh to the multi-view photogrammetry observations while maintaining consistent topology by Laplacian deformation. Optical flow-based refinement is also employed for temporal stability. 

We apply a rigid Procrustes alignment to eliminate the head poses and align each reconstructed mesh to a canonical coordinate frame shared across all timesteps (see details in supplementary materials) to ensure that the decoder focuses on predicting non-rigid, expression-related deformations. For each aligned mesh $M^{(t)}$, we then compute a corresponding 3D position map $P^{(t)} \in \mathbb{R}^{256\times256\times3}$, where each pixel encodes the $(x, y, z)$ coordinates of the facial surface in UV space. This representation preserves geometric continuity while enabling dense supervision for EEG decoding. Each position map is temporally aligned with its corresponding EEG window described in~\cref{par:EEG_preprocessing}, forming paired samples $(E^{(t)}, P^{(t)})$ for training the EEG-to-face decoder in~\cref{sec:conformer}.

\paragraph{\bf{Neural Avatar Preprocessing.}}
We convert the predicted 3D position maps $\hat{P}^{(t)}$ into meshes by sampling vertex coordinates $\hat{\mathbf{v}}^{(t)} \in \mathbb{R}^{V \times 3}$ to form meshes $\hat{M}^{(t)}$ according to the template UV layout. While the shared topology provides spatial consistency, the temporal smoothness of the reconstructed sequence primarily came from the overlapping EEG windows used during decoding, which yields gradually varying position maps over time. To prepare training data for GaussianAvatars, each mesh $M^{(t)}$ is projected onto the corresponding multi-view images to generate binary foreground masks, which are used to extract white-background facial crops $\{I^{(t)}_c\}$. The synchronized meshes $\{\mathbf{v}^{(t)}, \mathbf{f}\}$ and masked images $\{I^{(t)}_c\}$ are then used to train our modified GaussianAvatars model. 

\subsection{Decoding 3D Position Maps from EEG}
\label{sec:conformer}
To map EEG signals to dense 3D position maps, we first encode the preprocessed signals using a hybrid Convolution–Transformer architecture inspired by EEG-Conformer~\cite{song2023eeg}. Each EEG slice $E^{(t)} \in \mathbb{R}^{W \times C}$ is reshaped into a single-channel tensor of size $1 \times C \times W$, which can be interpreted as a spatiotemporal image capturing both temporal dynamics and inter-channel structure. Following prior work~\cite{Schirrmeister2017, RIYAD2021109037}, we apply two sequential convolutional layers: the first along the temporal axis to model temporal patterns, and the second across channels to capture spatial correlations between electrodes. The resulting features pass through batch normalization, an ELU activation, and an average-pooling layer that reduces temporal redundancy and promotes more stable, emotion-relevant representations.

The resulting feature sequence is passed through a transformer encoder that models long-range temporal dependencies within the EEG window and outputs a compact latent embedding $\mathbf{z}^{(t)}$. The position map provides a UV-parameterized mapping from surface geometry to a regular 2D grid in texture space, enabling efficient processing with standard 2D convolutional networks and image decoders. So, we employ a latent variational autoencoder(VAE) to generate the dense 3D position map $\hat{P}^{(t)}$, bridging the spatiotemporal EEG domain with the spatial structure of the human face.

A binary mask $m$ restricts supervision to the inner-face region, where facial expression dynamics occur, encouraging the decoder to focus on expressive geometry rather than rigid head motion. To further enhance surface smoothness and continuity, we include a self-supervised Laplacian regularization term (details in the Supplementary Materials). The overall position map loss is defined as:
\[
\mathcal{L_{\mathrm{pm}}} = \lambda_{\mathrm{rec}}  \mathcal{L}_{\mathrm{rec}} + \lambda_{\mathrm{smooth}} \mathcal{L}_{\mathrm{smooth}},
\]
where $\mathcal{L}_{\mathrm{rec}}$ denotes the per-pixel mean squared error (MSE), and $\mathcal{L}_{\mathrm{smooth}}$ represents the Laplacian smoothing loss applied to $\hat{P}$ within the masked region.

\begin{figure}[htbp]
    \centering
    \setlength{\belowcaptionskip}{-10pt}
    \includegraphics[width=0.5\linewidth]{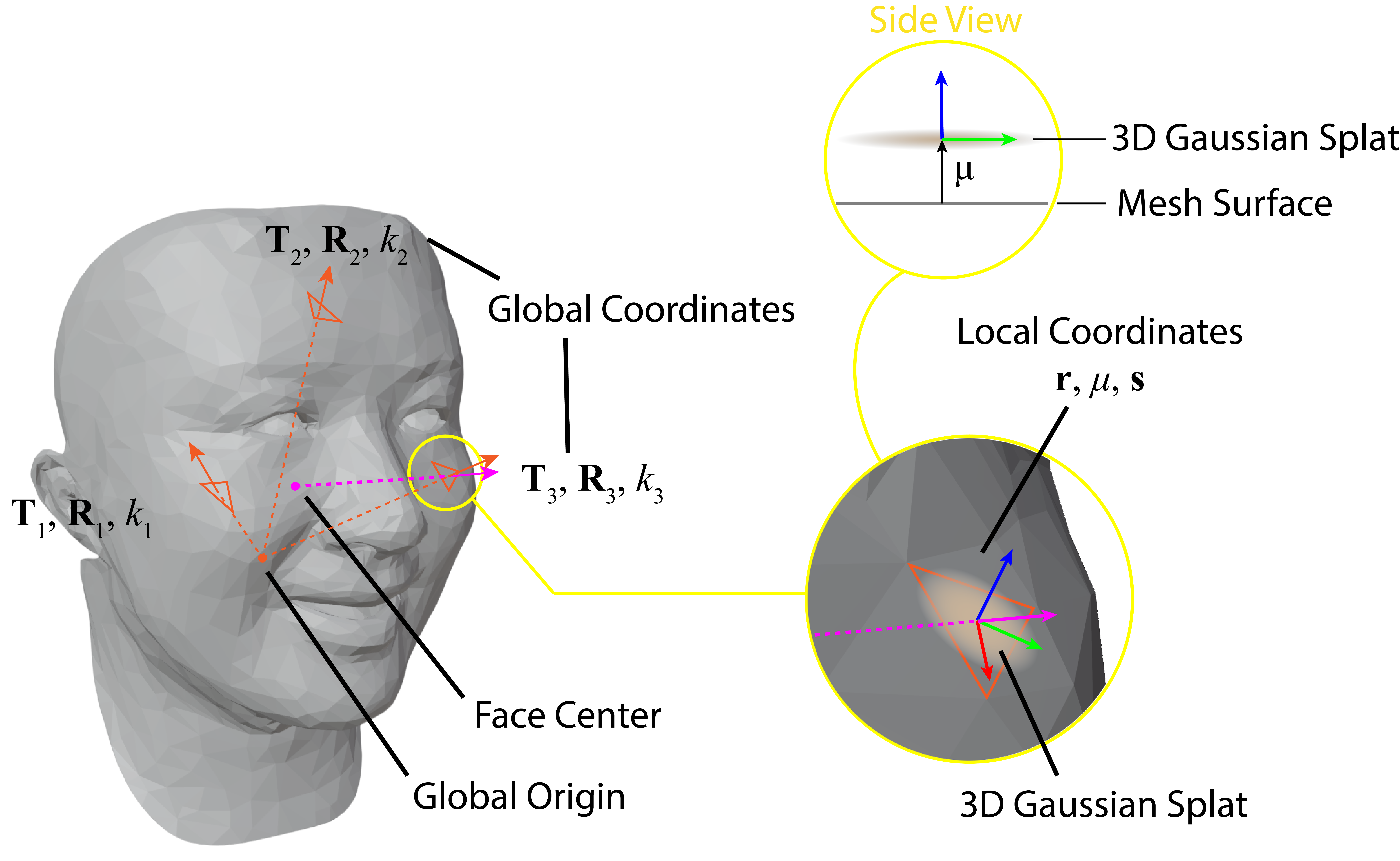}
    \caption{\textbf{Illustration of the Gaussian Binding Strategy.}~Each ellipsoidal Gaussian splat is defined by its local rotation $\mathbf{r}$, position $\mathbf{\mu}$, and scale $\mathbf{s}$ relative to the face center with global transform $\mathbf{T}$ and $\mathbf{R}$. During animation, $\mathbf{T}$ and $\mathbf{R}$ are updated over time according to the predicted facial geometry, while $\mathbf{r}$, $\mathbf{\mu}$, and $\mathbf{s}$ remain fixed as learnable parameters. The mesh is decimated for clarity in visualization.}
    \label{fig:gsavatar-illustration}
\end{figure}

\subsection{Avatar Rendering with 3D Gaussian Splatting}
\label{sec:method:modeling}
 We adopt 3DGS for avatar head rendering, replacing traditional texture-based methods with a view-dependent representation that captures realistic lighting via spherical harmonics. In this framework, a scene is modeled as a set of ellipsoidal Gaussian splats parameterized by a covariance matrix $\Sigma$ and center $\mathbf{\mu}$:
\[
G(\mathbf{x}) = \exp\!\left(-\tfrac{1}{2}(\mathbf{x} - \mathbf{\mu})^{\!T}\Sigma^{-1}(\mathbf{x} - \mathbf{\mu})\right),
\]
where $\Sigma$ is defined by rotation $\mathbf{r} \in \mathbb{R}^{3\times3}$, scale $\mathbf{s} \in \mathbb{R}^3$, and position $\mathbf{\mu} \in \mathbb{R}^3$. Each Gaussian’s color $\mathbf{c}$ is modeled using 3rd-order spherical harmonics, and the final color is rendered by alpha compositing overlapping splats:
\[
\mathbf{C} = \sum_{i=1}\mathbf{c}_i\alpha_i'\!\!\prod_{j<i}(1-\alpha_j')
\]
At initialization, each triangular face of the mesh is assigned one Gaussian with local parameters $(\mathbf{r}, \mathbf{\mu}, \mathbf{s})$, transformed into global coordinates during rendering as:
\begin{equation}
\mathbf{r}'=\mathbf{Rr},\quad \mathbf{\mu}'=k\mathbf{R\mu}+\mathbf{T},\quad \mathbf{s}'=k\mathbf{s},
\label{eqn:globtransform}
\end{equation}
where $\mathbf{R}$, $\mathbf{T}$, and $k$ denote the face’s rotation, translation, and scale, respectively. This provides a geometry-aware initialization for stable reconstruction (see \cref{fig:gsavatar-illustration}).

At the training phase, we do not compute the face geometry from the tracked FLAME parameters as in \cite{qian2024gaussianavatarsphotorealisticheadavatars}. Instead, we obtain directly the topology-consistent and facial geometry from our face tracking (see details in \cref{sec:preprocessing}), since it offers greater flexibility and thus fits closer to the actual vibrant facial expressions. Our face tracking pipeline also improves upon the baseline in our scenario with occlusion and extreme expressions, demonstrated in \cref{sec:comparison}. For each iteration, we sample a triplet of (mesh $M$, camera $p$, image $I$) uniformly at random, and render the Gaussian splats, each with the rigid transformation defined in~\cref{eqn:globtransform} based on $M$. The properties are optimized with loss \begin{equation}
    \mathcal{L}=(1-\lambda)\mathcal{L}_1 + \lambda\mathcal{L}_\text{D-SSIM} + \lambda_\text{pos}\mathcal{L}_\text{pos}+\lambda_\text{scale}\mathcal{L}_\text{scale},
\end{equation}
where $\mathcal{L}_1$ and $\mathcal{L}_\text{D-SSIM}$ compare the difference in the rendered images, $\mathcal{L}_\text{pos}=\|\max(\mathbf{\mu}, \epsilon_\text{pos})\|_2$ regularizes the spatial shift from the face center that the splat is bound to, and $\mathcal{L}_\text{scale}=\|\max(\mathbf{s}, \epsilon_\text{scale})\|_2$ regularizes scales of the Gaussians. During rendering, we move and render the optimized splats by the mesh deformed according to the position maps (see \cref{sec:conformer}) . For 3DGS training, we use the same hyperparameters as documented in \cite{qian2024gaussianavatarsphotorealisticheadavatars}.

\section{IMPLEMENTATION DETAILS}
\label{sec:implementation}

\paragraph{\bf{CNN-Transformer Encoder.}}
Each EEG input window is represented as a tensor of shape $(B, 1, 16, 375)$, where $16$ denotes the electrode channels and $375$ the temporal samples. The encoder begins with a temporal convolution with $40$ output channels and kernel size 40 to capture short-range temporal patterns. This is followed by a spatial convolution with kernel size 16, spanning all electrodes to model spatial correlations between channels. The resulting feature maps are processed by Batch Normalization, an ELU nonlinearity, and an average pooling layer to reduce temporal resolution while preserving salient dynamics. After pooling, the temporal dimension is reduced from $375$ to approximately $19$ feature patches. A $1 \times 1$ convolution then projects the feature maps to an embedding dimension of $E = 40$, forming a token sequence $\mathbf{z}_0^{(t)} \in \mathbb{R}^{B \times N \times E}$ with $N = 19$.

The token sequence $\mathbf{z}_0^{(t)}$ is then processed by a stack of six Transformer encoder layers. Each layer uses pre-normalization with LayerNorm, multi-head self-attention with $10$ heads, and a feed-forward network that expands the embedding dimension by a factor of $4$, followed by GELU activation and dropout ($p = 0.5$). Residual connections are applied to both attention and feed-forward sub-layers. The final output $\mathbf{z}^{(t)}$ encodes temporally integrated EEG features for time window $t$, which are passed to the 3D position map decoder.

\paragraph{\bf{Position Map Decoder.}}
To reconstruct dense 3D position maps from the EEG feature $\mathbf{z}^{(t)}$, we employ a latent variational autoencoder (VAE) as an image decoder. The decoder receives a flattened $z^{(t)}$ of size $B \times d_{\text{in}}$, which is projected through a sequence of fully connected layers into a latent feature map of dimension $B \times 4 \times 8 \times 8$. The latent tensor is then decoded through the VAE into a coarse 3-channel output, followed by two transposed convolution layers to produce a dense 3D position map $\hat{P} \in \mathbb{R}^{256 \times 256 \times 3}$, where each pixel encodes the $(x, y, z)$ coordinates of the reconstructed facial surface.

The decoder parameters are randomly initialized and trained jointly with the EEG transformer encoder to predict dense 3D position maps from EEG signals. The predicted face mesh is sampled from the resulting position map and used to drive the trained Gaussian splats attached to the faces to render the final facial image.

\section{EXPERIMENTS}
As a pioneering work, there is no direct baseline to compare against; we present visual results under different emotional stimuli of two subjects as our own qualitative results to assess system performance.

\paragraph{\bf {Data Split.}} Our data for evaluation consists of two types of trials: the emotion-specific ones (Angry, Disgust, Funny, Neutral, Sad) and the standalone evaluation trial \textit{EMOSTIM}. Each emotion trial consists of several stimulus videos presented sequentially to the subject, during which synchronized EEG and multi-view facial recordings are captured. For the emotion-specific trials, we use responses corresponding to the final 15\% of frames of each trial as the test set and the remaining frames for training, following standard signal processing practices to minimize statistical leakage. In addition, the standalone \textit{EMOSTIM} sequence, which contains a distinct set of mixed-emotion stimulus videos, is excluded from training and used solely to evaluate cross-trial generalization.

\begin{figure}[!htb]
\includegraphics[width=1.0\linewidth]{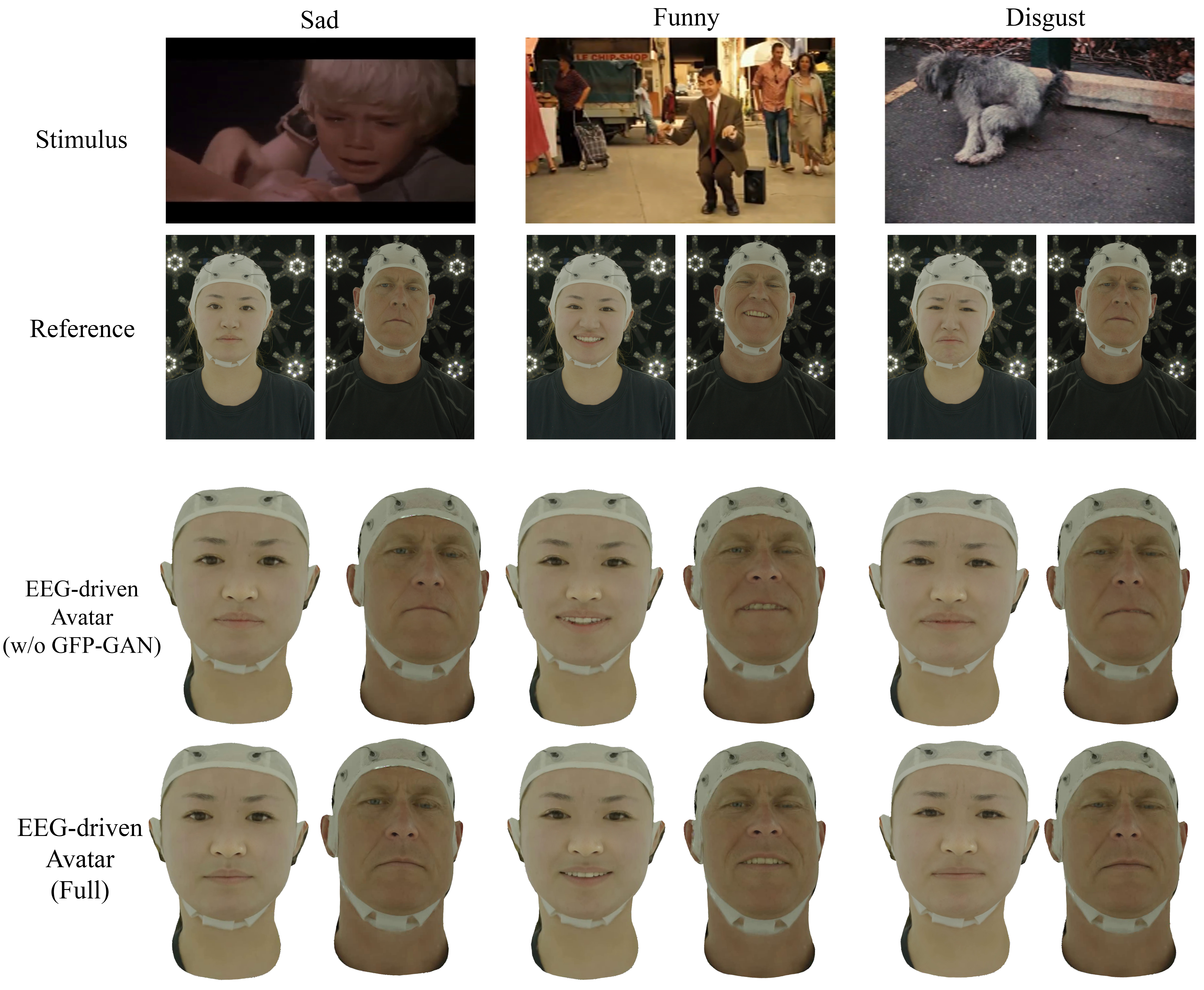}
\caption{\textbf{Qualitative Results of EEG-driven Facial Expression Synthesis}. For three trials (Sad, Funny, Disgust), we show the emotion-eliciting video frames from the testing set (first row), the corresponding captured facial expressions from two subjects (second row), the corresponding renders produced by our EEG-driven avatar (third row), and our final result produced with mouth-interior enhancement (fourth row). The reconstructed avatars reflect emotional responses (\eg frown for sadness, smiles for funny clips, brow/nose tightening for disgust) with subject-specific intensities and capture both neutral and strongly expressive frames. The GFP-GAN enhancement is applied only to the mouth interior and does not modify the emotion response (\cf the third and fourth row). Quantitative metrics can be found in supplementary materials.} 

\label{fig:qual_fig}
\end{figure}

\paragraph{\bf Qualitative Results.}
\cref{fig:qual_fig} shows responses of two subjects across three stimulus types. The EEG-driven avatars reproduce the facial expressions while preserving each subject's identity and adapting to subject-specific dynamics, such as smile intensity, indicating effective personalization rather than broad categorization. For avatar rendering, we further refine the mouth interior using a pretrained GFP-GAN~\cite{wang2021towards} model, following the post-processing strategies adopted in recent neural avatar works~\cite{li2023generalizable,tran2024voodoo}. The enhancement is strictly confined to the mouth interior, which is frequently occluded during the capture and thus hard to be modeled by GS-based avatar models, and does not tamper with the decoded emotion, as we show in the third and fourth columns of the figure. More results, including multi-view rendering and videos are provided in Supplementary Materials. 
\clearpage

\section{ABLATION STUDIES}
In this section, we conduct an ablation study to evaluate the impact of key components of our method and justify our design choices.

\subsection{Comparison with GaussianAvatars}
\label{sec:comparison}

\begin{figure}[htb]
    \centering
    \includegraphics[width=0.6\linewidth]{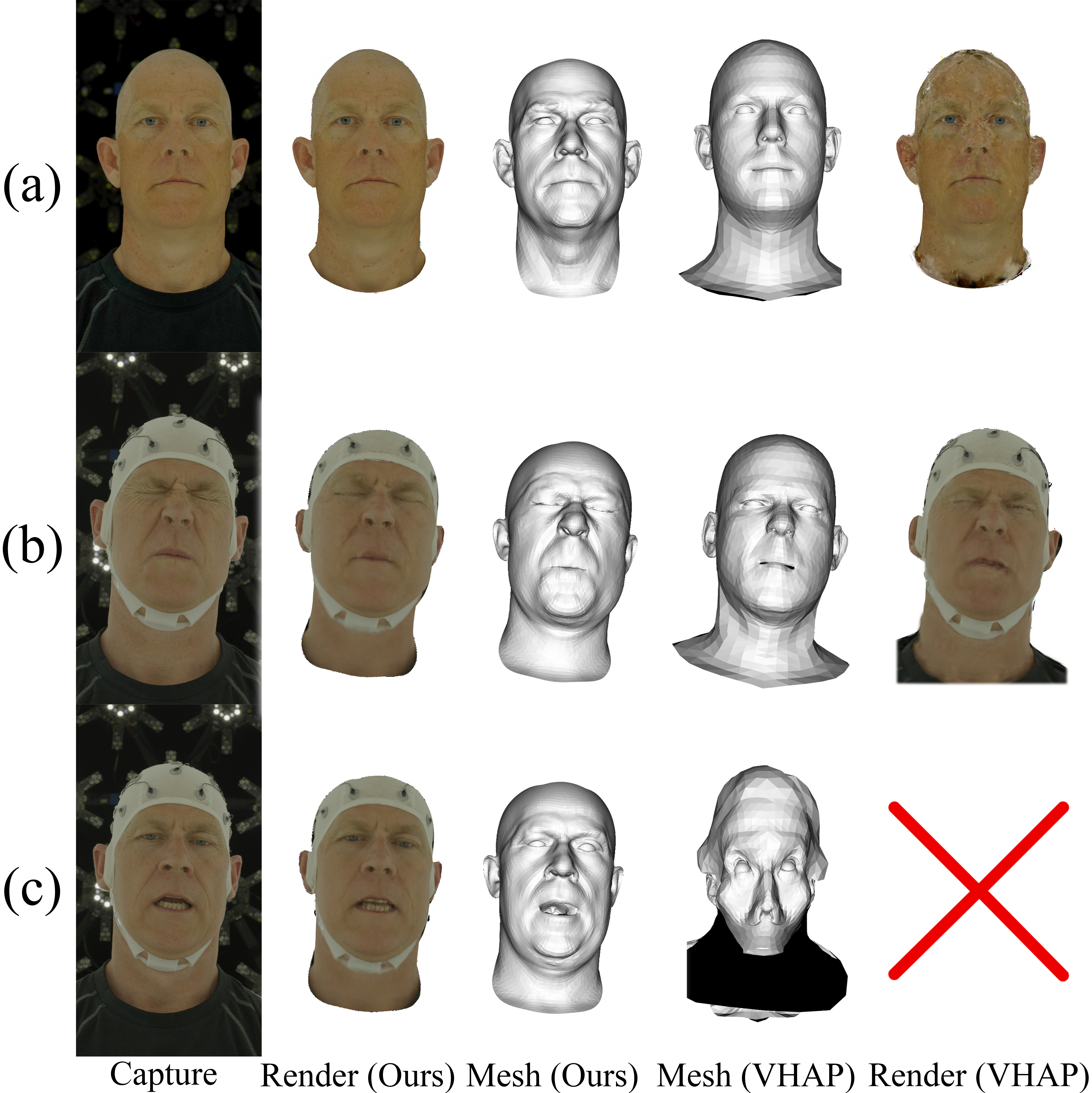}
    \caption{\textbf{We demonstrate that the face tracking pipeline of GaussianAvatars does not suit our scenario.} In (a), we show that in an occlusion-free scenario, the avatar rendered by our pipeline is similar to that of GaussianAvatars. However, in our complex capture setup with extreme expressions and headcap-induced occlusion, errors in landmark detection of the GaussianAvatars pipeline can lead to incorrect expression tracking (b) or even completely broken geometry (c). The red cross indicates that the model fails to produce a faithful render due to collapsed geometry.
    }
    \label{fig:comparison}
\end{figure}

The GaussianAvatars pipeline is optimal in the common scenario where the expressions are mild and the majority of the face is clear. However, in our setting, to capture synchronized image and EEG data, it is necessary for the subjects to wear a headcap with sensors, which may confuse common facial landmark decoders, such as the VHAP library~\cite{qian2024vhap} used in their pipeline; the extreme expressions, triggered by the emotion-eliciting videos we use, are often beyond the expressivity of FLAME parameters~\cite{FLAME2017}. Therefore, as demonstrated by \cref{fig:comparison}, our ablation study shows that the standard GaussianAvatar pipeline does not generalize to our data capture setup, and it may predict inaccurate expressions (b) or even fail catastrophically (c), whereas our method maintains robustness in the complex capture settings. 

\subsection{Why Position Maps Instead of Blendshapes?}
\label{sec:bs_pm_comparison}
\begin{figure}[htb]
    \centering
    \includegraphics[width=0.8\linewidth]{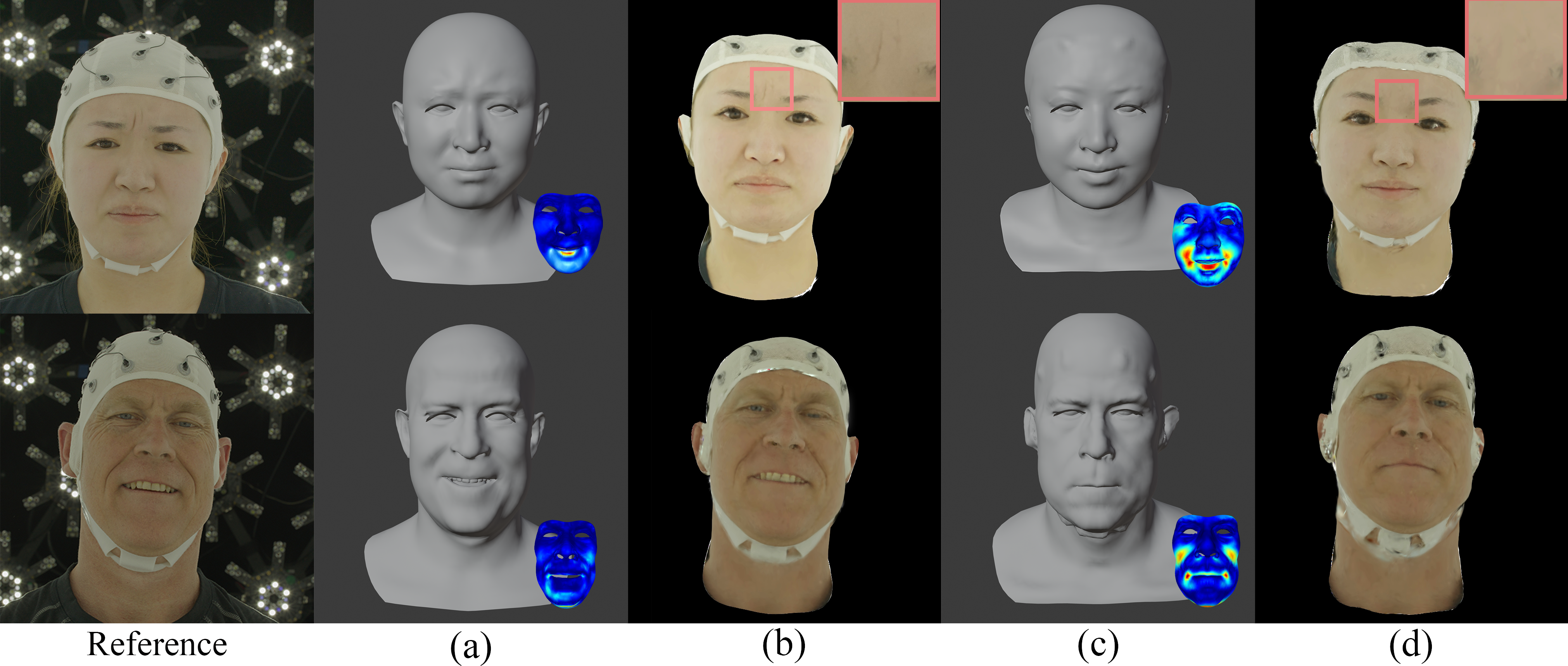}
    \caption{\textbf{Qualitative comparison between position map–driven and blendshape-driven avatars.~} Subject 1 (top) and Subject 2 (bottom). From left to right: Reference image, (a) position map–driven avatar mesh, (b) position map–driven avatar render, (c) blendshape-driven avatar mesh, and (d) blendshape-driven avatar mesh.}
    \label{fig:bs_pm_comparison}
\end{figure}

\begin{table}[t]
\centering
\footnotesize
\setlength{\tabcolsep}{3pt}
\renewcommand{\arraystretch}{1.15}
\resizebox{1.0\linewidth}{!}{
\begin{tabular}{lcccccc|cccccc}
\toprule
& \multicolumn{6}{c}{\textbf{Position Map-driven Avatar}} 
& \multicolumn{6}{c}{\textbf{Blendshape-driven Avatar}} \\
\cmidrule(lr){2-7} \cmidrule(lr){8-13}

& \multicolumn{3}{c}{\textbf{Subject 1}} 
& \multicolumn{3}{c}{\textbf{Subject 2}}
& \multicolumn{3}{c}{\textbf{Subject 1}} 
& \multicolumn{3}{c}{\textbf{Subject 2}} \\

\cmidrule(lr){2-4} \cmidrule(lr){5-7} 
\cmidrule(lr){8-10} \cmidrule(lr){11-13}

\textbf{Trial} 
& Mean & $<1$mm(\%) & $<3$mm(\%)
& Mean & $<1$mm(\%) & $<3$mm(\%)
& Mean & $<1$mm(\%) & $<3$mm(\%)
& Mean & $<1$mm(\%) & $<3$mm(\%) \\

\midrule
Angry   & 0.4982 & 87.37 & 99.67 & 0.5534 & 84.28 & 99.64 & 1.0792 & 60.76 & 93.94 & 1.6492 & 42.79 & 83.92 \\
Disgust & 1.0238 & 61.89 & 95.12 & 0.6347 & 80.57 & 98.78 & 1.7680 & 42.03 & 81.85 & 1.3410 & 60.35 & 85.66 \\
Funny   & 0.9614 & 69.12 & 92.82 & 0.7432 & 78.24 & 96.53 & 1.3385 & 56.56 & 87.08 & 1.6195 & 48.93 & 83.58 \\
Neutral & 0.4028 & 92.85 & 100.0 & 0.3245 & 95.84 & 99.92 & 0.3270 & 94.99 & 99.95 & 1.3065 & 52.91 & 89.98 \\
Sad     & 0.3571 & 94.60 & 99.98 & 0.3112 & 96.60 & 99.55 & 0.5391 & 84.95 & 99.59 & 1.3134 & 56.22 & 88.15 \\

\midrule
EMOSTIM
& 0.5158 & 86.76 & 99.16 & 0.7077 & 78.88 & 97.55 
& 0.6838 & 78.78 & 97.78 & 1.5078 & 52.22 & 86.28 \\

\bottomrule
\end{tabular}
}

\caption{\textbf{Quantitative evaluation using point-to-surface distance within the masked facial region.} Errors are reported in millimeters (mm), including the mean distance and the percentage of vertices below 1 mm and 3 mm. The EMOSTIM trial is used as a leave-one-out trial to evaluate cross-trial generalization.}
\label{tab:pm_metrics}
\end{table}

Using blendshapes presents two key limitations. First, mapping EEG signals to a fixed set of 52 coefficients produces a sparse and less expressive representation, making it difficult to capture subtle and nuanced facial expressions. Second, standard blendshape models contain relatively few vertices ($\sim$5,000 in FLAME), which limits their ability to represent mid-frequency geometric details such as wrinkles. In contrast, our approach directly maps EEG signals to position maps derived from ground-truth multi-view stereo reconstructions. A ${256 \times 256}$ position map ($\sim$65k vertices) preserves mid-frequency details that blendshape models cannot capture. By avoiding the two-step dimensionality reduction inherent in blendshape-based approaches, the position-map representation better preserves the underlying geometric fidelity.

To evaluate the practical impact of these representations, we compare avatars driven by predicted blendshapes against those driven by predicted position maps. As shown in \cref{fig:bs_pm_comparison}, the blendshape-driven reconstructions produce plausible facial motion but appear overly smoothed and exhibit reduced expressive amplitude-a consequence of the limited dimensionality of the blendshape basis. Conversely, the position map representation preserves richer geometric detail, allowing wrinkles and localized facial structures to emerge more naturally in the rendered avatars.

\cref{tab:pm_metrics} further quantifies this difference. Across all trials except for Neutral, the position map–driven avatar consistently achieves lower geometric error. These results suggest that the compact blendshape basis acts as a strong bottleneck when mapping noisy signals, often leading the model toward ``averaged-out'' predictions. The higher-resolution geometry of the position map yields more accurate surface reconstruction; specifically, the percentage of vertices with error below 1 mm is consistently higher for the position map–driven avatar. Ultimately, these results indicate that the dense position map representation not only produces visually richer animations but also yields more accurate geometric reconstructions compared to the low-dimensional blendshape representation in an EEG-driven context.

\section{CONCLUSIONS AND FUTURE WORK}
\label{sec:conclusions}

We introduce Mind-to-Face, the first framework that maps non-invasive EEG signals directly to photorealistic 3D facial expressions. By decoding EEG into dense 3D position maps and rendering them with 3D Gaussian Splatting, our approach enables subject-specific facial animation driven entirely by neural activity. A synchronized EEG–video capture setup provides the precise alignment needed for supervised learning, establishing a unified pipeline for neural-driven avatars and new opportunities in affective computing, telepresence, and brain–computer interfaces. In future work, we plan to expand the dataset with more subjects and diverse stimuli to improve robustness and cross-subject generalization, and incorporate stimulus context during decoding to better disentangle perceptual and affective signals for more accurate facial synthesis.


\bibliographystyle{splncs04}
\bibliography{main}

@String(CVPR  = {IEEE Conf. Comput. Vis. Pattern Recog.})

@String(ECCV  = {Eur. Conf. Comput. Vis.})

@String(TOG   = {ACM Trans. Graph.})

@String(CVPR  = {CVPR})

@String(ECCV  = {ECCV})

@String(TOG   = {ACM TOG})

@article{kerbl20233dgaussiansplattingrealtime,
author = {Kerbl, Bernhard and Kopanas, Georgios and Leimkuehler, Thomas and Drettakis, George},
title = {3D Gaussian Splatting for Real-Time Radiance Field Rendering},
year = {2023},
issue_date = {August 2023},
publisher = {Association for Computing Machinery},
address = {New York, NY, USA},
volume = {42},
number = {4},
issn = {0730-0301},
url = {https://doi.org/10.1145/3592433},
doi = {10.1145/3592433},
journal = {ACM Trans. Graph.},
month = jul,
articleno = {139},
numpages = {14}
}

@INPROCEEDINGS{10405666,
  author={Panachakel, Jerrin Thomas and H, Ranjana and K, Sana Parveen and Sidharth, Sidharth and Samuel, Ashish Abraham},
  booktitle={2023 IEEE International Conference on Metrology for eXtended Reality, Artificial Intelligence and Neural Engineering (MetroXRAINE)}, 
  title={CSP- LSTM Based Emotion Recognition from EEG Signals}, 
  year={2023},
  volume={},
  number={},
  pages={289-294},
  doi={10.1109/MetroXRAINE58569.2023.10405666}}

@article{sun2023survey, 
  year     = {2023}, 
  title    = {Survey on the research direction of {EEG}-based signal processing}, 
  author   = {Sun, Congzhong and Mou, Chaozhou}, 
  journal  = {Frontiers in Neuroscience}, 
  issn     = {1662-4548}, 
  doi      = {10.3389/fnins.2023.1203059}, 
  pmid     = {37521708}, 
  pmcid    = {{PMC}10372445}, 
  pages    = {1203059}, 
  volume   = {17}
}

@ARTICLE{Zheng2018,
  author={Zheng, Wei-Long and Liu, Wei and Lu, Yifei and Lu, Bao-Liang and Cichocki, Andrzej},
  journal={IEEE Transactions on Cybernetics}, 
  title={EmotionMeter: A Multimodal Framework for Recognizing Human Emotions}, 
  year={2018},
  volume={49},
  number={3},
  pages={1110-1122},
  doi={10.1109/TCYB.2018.2797176}}

@article{Zheng2017,
doi = {10.1088/1741-2552/aa5a98},
url = {https://dx.doi.org/10.1088/1741-2552/aa5a98},
year = {2017},
month = {feb},
publisher = {IOP Publishing},
volume = {14},
number = {2},
pages = {026017},
author = {Zheng, Wei-Long and Lu, Bao-Liang},
title = {A multimodal approach to estimating vigilance using EEG and forehead EOG},
journal = {Journal of Neural Engineering}
}

@inproceedings{Duan2013,
author = {Duan, Ruo-Nan and Zhu, Jia-Yi and Lu, Bao-Liang},
year = {2013},
month = {11},
pages = {81-84},
title = {Differential entropy feature for EEG-based emotion classification},
booktitle = {International IEEE/EMBS Conference on Neural Engineering},
doi = {10.1109/NER.2013.6695876}
}

@MISC{Israel2021,
  title     = "Open Library for Affective Videos ({OpenLAV})",
  author    = "Israel, Laura and Paukner, Philipp and Schiestel, Lena and
               Diepold, Klaus and Sch{\"o}nbrodt, Felix",
  publisher = "PsychArchives",
  year      =  2021
}

@article{song2023eeg, 
  year     = {2023}, 
  title    = {{EEG} Conformer: Convolutional Transformer for {EEG} Decoding and Visualization}, 
  author   = {Song, Yonghao and Zheng, Qingqing and Liu, Bingchuan and Gao, Xiaorong}, 
  journal  = {{IEEE} Transactions on Neural Systems and Rehabilitation Engineering}, 
  issn     = {1534-4320}, 
  doi      = {10.1109/tnsre.2022.3230250}, 
  pmid     = {37015413}, 
  pages    = {710--719}, 
  volume   = {31}
}

@article{ekman1992argument, 
  year     = {1992}, 
  title    = {An argument for basic emotions}, 
  author   = {Ekman, Paul}, 
  journal  = {Cognition and Emotion}, 
  issn     = {0269-9931}, 
  doi      = {10.1080/02699939208411068}, 
  pages    = {169--200}, 
  number   = {3-4}, 
  volume   = {6}
}

@inproceedings{zielonka2025drivable3dgaussianavatars,
  title        = {Drivable 3D Gaussian Avatars},
  author       = {Wojciech Zielonka and Timur Bagautdinov and Shunsuke Saito and Michael Zollhöfer and Justus Thies and Javier Romero},
  booktitle    = {I3DV},
  month        = {March},
  year         = {2025}
}

@inproceedings{moon2024expressivewholebody3dgaussian,
      title={Expressive Whole-Body 3D Gaussian Avatar}, 
          author={Gyeongsik Moon and Takaaki Shiratori and Shunsuke Saito},
    booktitle={ECCV},
    year={2024} 
}

@inproceedings{yuan2023gavatar,
  title={GAvatar: Animatable 3D Gaussian Avatars with Implicit Mesh Learning},
  author={Yuan, Ye and Li, Xueting and Huang, Yangyi and De Mello, Shalini and Nagano, Koki and Kautz, Jan and Iqbal, Umar},
  booktitle={Proceedings of the IEEE/CVF conference on computer vision and pattern recognition},
  pages={896--905},
  year={2024}
}

@inproceedings{tang2025gafgaussianavatarreconstruction,
      title={GAF: Gaussian Avatar Reconstruction from Monocular Videos via Multi-view Diffusion}, 
      author={Tang, Jiapeng and Davoli, Davide and Kirschstein, Tobias and Schoneveld, Liam and Niessner, Matthias},
      booktitle={Proceedings of the Computer Vision and Pattern Recognition Conference},
      pages={5546--5558},
      year={2025}
    }

@inproceedings{hu2024gaussianavatar,
        title={GaussianAvatar: Towards Realistic Human Avatar Modeling from a Single Video via Animatable 3D Gaussians},
        author={Hu, Liangxiao and Zhang, Hongwen and Zhang, Yuxiang and Zhou, Boyao and Liu, Boning and Zhang, Shengping and Nie, Liqiang},
        booktitle={IEEE/CVF Conference on Computer Vision and Pattern Recognition (CVPR)},
        year={2024}
}

@article{FLAME2017, 
  title = {Learning a model of facial shape and expression from {4D} scans}, 
  author = {Li, Tianye and Bolkart, Timo and Black, Michael. J. and Li, Hao and Romero, Javier}, 
  journal = {ACM Transactions on Graphics, (Proc. SIGGRAPH Asia)}, 
  volume = {36}, 
  number = {6}, 
  year = {2017}, 
  pages = {194:1--194:17},
  url = {https://doi.org/10.1145/3130800.3130813} 
}

@misc{qian2024vhap,
  title={VHAP: Versatile Head Alignment with Adaptive Appearance Priors},
  author={Qian, Shenhan},
  year={2024},
  month={sep},
  doi={10.5281/zenodo.14988309},
  url={https://github.com/ShenhanQian/VHAP}
}

@inproceedings{grassal2022neuralheadavatarsmonocular,
      title={Neural Head Avatars from Monocular RGB Videos}, 
       author={Grassal, Philip-William and Prinzler, Malte and Leistner, Titus and Rother, Carsten and Nie{\ss}ner, Matthias and Thies, Justus},
      booktitle={Proceedings of the IEEE/CVF conference on computer vision and pattern recognition},
      pages={18653--18664},
      year={2022}

}

@inproceedings{zheng2022imavatar,
  title={{I} {M} {Avatar}: Implicit Morphable Head Avatars from Videos},
  author={Zheng, Yufeng and Abrevaya, Victoria Fernández and Bühler, Marcel C. and Chen, Xu and Black, Michael J. and Hilliges, Otmar},
  booktitle = {Computer Vision and Pattern Recognition (CVPR)},
  year = {2022}
}

@article{mildenhall2020nerfrepresentingscenesneural,
author = {Mildenhall, Ben and Srinivasan, Pratul P. and Tancik, Matthew and Barron, Jonathan T. and Ramamoorthi, Ravi and Ng, Ren},
title = {NeRF: representing scenes as neural radiance fields for view synthesis},
year = {2021},
issue_date = {January 2022},
publisher = {Association for Computing Machinery},
address = {New York, NY, USA},
volume = {65},
number = {1},
issn = {0001-0782},
url = {https://doi.org/10.1145/3503250},
doi = {10.1145/3503250},
journal = {Commun. ACM},
month = dec,
pages = {99–106},
numpages = {8}
}

@InProceedings{Gafni_2021_CVPR,
    author    = {Gafni, Guy and Thies, Justus and Zollh{\"o}fer, Michael and Nie{\ss}ner, Matthias},
    title     = {Dynamic Neural Radiance Fields for Monocular 4D Facial Avatar Reconstruction},
    booktitle = {Proceedings of the IEEE/CVF Conference on Computer Vision and Pattern Recognition (CVPR)},
    month     = {June},
    year      = {2021},
    pages     = {8649-8658}
}

@article{kim2018deep,
  title     = {Deep Video Portraits},
  author    = {Kim, Hyeongwoo and Garrido, Pablo and Tewari, Ayush and Xu, Weipeng and Thies, Justus and 
               Nie{\ss}ner, Matthias and P{\'e}rez, Patrick and Richardt, Christian and Zollh{\"o}fer, Michael and Theobalt, Christian},
  journal   = {ACM Transactions on Graphics (TOG)},
  volume    = {37},
  number    = {4},
  pages     = {163},
  year      = {2018}
}

@inproceedings{Xu2023, 
      year = {2023}, 
      title = {{LatentAvatar: Learning Latent Expression Code for Expressive Neural Head Avatar}}, 
      author={Xu, Yuelang and Zhang, Hongwen and Wang, Lizhen and Zhao, Xiaochen and Huang, Han and Qi, Guojun and Liu, Yebin},
      booktitle={ACM SIGGRAPH 2023 Conference Proceedings},
      pages={1--10}
}

@article{Suwajanakorn2017,
author = {Suwajanakorn, Supasorn and Seitz, Steven M. and Kemelmacher-Shlizerman, Ira},
title = {Synthesizing Obama: learning lip sync from audio},
year = {2017},
issue_date = {August 2017},
publisher = {Association for Computing Machinery},
address = {New York, NY, USA},
volume = {36},
number = {4},
issn = {0730-0301},
url = {https://doi.org/10.1145/3072959.3073640},
doi = {10.1145/3072959.3073640},
journal = {ACM Trans. Graph.},
month = jul,
articleno = {95},
numpages = {13}
}

@article{Thies2019, 
  year = {2019}, 
  title = {{Deferred Neural Rendering: Image Synthesis using Neural Textures}}, 
  author={Thies, Justus and Zollh{\"o}fer, Michael and Nie{\ss}ner, Matthias},
  journal={Acm Transactions on Graphics (TOG)},
  volume={38},
  number={4},
  pages={1--12},
  publisher={ACM New York, NY, USA}
}

@inproceedings{Chan2018, 
  title = {{Everybody Dance Now}}, 
  author={Chan, Caroline and Ginosar, Shiry and Zhou, Tinghui and Efros, Alexei A},
  booktitle={Proceedings of the IEEE/CVF international conference on computer vision},
  pages={5933--5942},
  year={2019}
}

@inproceedings{Thies2016, 
  year = {2016}, 
  title = {{Face2Face: Real-time Face Capture and Reenactment of RGB Videos}}, 
  author={Thies, Justus and Zollhofer, Michael and Stamminger, Marc and Theobalt, Christian and Nie{\ss}ner, Matthias},
  booktitle={Proceedings of the IEEE conference on computer vision and pattern recognition},
  pages={2387--2395},
}

@article{Sollfrank2021, 
year = {2021}, 
title = {{The Effects of Dynamic and Static Emotional Facial Expressions of Humans and Their Avatars on the EEG: An ERP and ERD/ERS Study}}, 
author = {Sollfrank, Teresa and Kohnen, Oona and Hilfiker, Peter and Kegel, Lorena C. and Jokeit, Hennric and Brugger, Peter and Loertscher, Miriam L. and Rey, Anton and Mersch, Dieter and Sternagel, Joerg and Weber, Michel and Grunwald, Thomas}, 
journal = {Frontiers in Neuroscience}, 
issn = {1662-4548}, 
doi = {10.3389/fnins.2021.651044}, 
pmid = {33967681}, 
pmcid = {PMC8100234}, 
pages = {651044}, 
volume = {15}
}

@inproceedings{Liu2010,
author = {Liu, Yisi and Sourina, Olga and Nguyen, Minh Khoa},
title = {Real-Time EEG-Based Human Emotion Recognition and Visualization},
year = {2010},
isbn = {9780769542157},
publisher = {IEEE Computer Society},
address = {USA},
url = {https://doi.org/10.1109/CW.2010.37},
doi = {10.1109/CW.2010.37},
booktitle = {Proceedings of the 2010 International Conference on Cyberworlds},
pages = {262–269},
numpages = {8},
series = {CW '10}
}

@article{russell1980circumplex, 
  year     = {1980}, 
  title    = {A circumplex model of affect}, 
  author   = {Russell, James A.}, 
  journal  = {Journal of Personality and Social Psychology}, 
  issn     = {0022-3514}, 
  doi      = {10.1037/h0077714}, 
  pages    = {1161--1178}, 
  number   = {6}, 
  volume   = {39}
}

@article{li2015,
author = {Li, Hao and Trutoiu, Laura and Olszewski, Kyle and Wei, Lingyu and Trutna, Tristan and Hsieh, Pei-Lun and Nicholls, Aaron and Ma, Chongyang},
year = {2015},
month = {08},
pages = {},
title = {Facial-Performance-Sensing Head-Mounted Display},
volume = {34},
journal = {ACM Transactions on Graphics},
doi = {10.1145/2766939}
}

@article{zheng2015SEED, 
  year     = {2015}, 
  title    = {Investigating Critical Frequency Bands and Channels for {EEG}-Based Emotion Recognition with Deep Neural Networks}, 
  author   = {Zheng, Wei-Long and Lu, Bao-Liang}, 
  journal  = {{IEEE} Transactions on Autonomous Mental Development}, 
  issn     = {1943-0604}, 
  doi      = {10.1109/tamd.2015.2431497}, 
  pages    = {162--175}, 
  number   = {3}, 
  volume   = {7}
}

@inproceedings{Nadeem24, 
  year = {2024}, 
  title = {{EVOKE: Emotion Enabled Virtual Avatar Mapping Using Optimized Knowledge Distillation}}, 
  author={Nadeem, Maryam and Imam, Raza and Al-Refai, Rouqaiah and Chkir, Meriem and Hoda, Mohamad and El Saddik, Abdulmotaleb},
  booktitle={2024 IEEE International Conference on Consumer Electronics (ICCE)},
  pages={1--6},
  organization={IEEE}
}

@article{liu2022rapid,
  title={Rapid face asset acquisition with recurrent feature alignment},
  author={Liu, Shichen and Cai, Yunxuan and Chen, Haiwei and Zhou, Yichao and Zhao, Yajie},
  journal={ACM Transactions on Graphics (TOG)},
  volume={41},
  number={6},
  pages={1--17},
  year={2022},
  publisher={ACM New York, NY, USA}
}

@article{Ameri2018, 
year = {2018}, 
title = {{Imperceptible electrooculography graphene sensor system for human–robot interface}}, 
author = {Ameri, Shideh Kabiri and Kim, Myungsoo and Kuang, Irene Agnes and Perera, Withanage K and Alshiekh, Mohammed and Jeong, Hyoyoung and Topcu, Ufuk and Akinwande, Deji and Lu, Nanshu}, 
journal = {npj 2D Materials and Applications}, 
doi = {10.1038/s41699-018-0064-4}, 
pages = {19}, 
number = {1}, 
volume = {2}
}

@article{Jiang2019, 
year = {2019}, 
title = {{BrainNet: A Multi-Person Brain-to-Brain Interface for Direct Collaboration Between Brains}}, 
author = {Jiang, Linxing and Stocco, Andrea and Losey, Darby M and Abernethy, Justin A and Prat, Chantel S and Rao, Rajesh P N}, 
journal = {Scientific Reports}, 
doi = {10.1038/s41598-019-41895-7}, 
pages = {6115}, 
number = {1}, 
volume = {9}
}

@article{song2021transformerbased, 
  year     = {2021}, 
  title    = {Transformer-based Spatial-Temporal Feature Learning for {EEG} Decoding}, 
   author={Song, Yonghao and Jia, Xueyu and Yang, Lie and Xie, Longhan},
  journal={arXiv preprint arXiv:2106.11170},
}

@article{Schirrmeister2017,
author = {Schirrmeister, Robin Tibor and Springenberg, Jost Tobias and Fiederer, Lukas Dominique Josef and Glasstetter, Martin and Eggensperger, Katharina and Tangermann, Michael and Hutter, Frank and Burgard, Wolfram and Ball, Tonio},
title = {Deep learning with convolutional neural networks for EEG decoding and visualization},
journal = {Human Brain Mapping},
volume = {38},
number = {11},
pages = {5391-5420},
doi = {https://doi.org/10.1002/hbm.23730},
url = {https://onlinelibrary.wiley.com/doi/abs/10.1002/hbm.23730},
eprint = {https://onlinelibrary.wiley.com/doi/pdf/10.1002/hbm.23730},
year = {2017}
}

@article{RIYAD2021109037,
title = {MI-EEGNET: A novel convolutional neural network for motor imagery classification},
journal = {Journal of Neuroscience Methods},
volume = {353},
pages = {109037},
year = {2021},
issn = {0165-0270},
doi = {https://doi.org/10.1016/j.jneumeth.2020.109037},
url = {https://www.sciencedirect.com/science/article/pii/S016502702030460X},
author = {Mouad Riyad and Mohammed Khalil and Abdellah Adib}
}

@article{ZHANG2023104157,
title = {EEG-based multi-frequency band functional connectivity analysis and the application of spatio-temporal features in emotion recognition},
journal = {Biomedical Signal Processing and Control},
volume = {79},
pages = {104157},
year = {2023},
issn = {1746-8094},
doi = {https://doi.org/10.1016/j.bspc.2022.104157},
url = {https://www.sciencedirect.com/science/article/pii/S1746809422006115},
author = {Yuchan Zhang and Guanghui Yan and Wenwen Chang and Wenqie Huang and Yueting Yuan}
}

@misc{postepski2024recurrentconvolutionalneuralnetworks,
      title={Recurrent and Convolutional Neural Networks in Classification of EEG Signal for Guided Imagery and Mental Workload Detection}, 
       author={Postepski, Filip and Wojcik, Grzegorz M and Wrobel, Krzysztof and Kawiak, Andrzej and Zemla, Katarzyna and Sedek, Grzegorz},
      journal={Scientific Reports},
      volume={15},
      number={1},
      pages={10521},
      year={2025},
      publisher={Nature Publishing Group UK London}
}

@article{Lawhern_2018,
   title={EEGNet: a compact convolutional neural network for EEG-based brain–computer interfaces},
   volume={15},
   ISSN={1741-2552},
   url={http://dx.doi.org/10.1088/1741-2552/aace8c},
   DOI={10.1088/1741-2552/aace8c},
   number={5},
   journal={Journal of Neural Engineering},
   publisher={IOP Publishing},
   author={Lawhern, Vernon J and Solon, Amelia J and Waytowich, Nicholas R and Gordon, Stephen M and Hung, Chou P and Lance, Brent J},
   year={2018},
   month=jul, pages={056013} }

@incollection{WOLPAW202015,
title = {Chapter 2 - Brain-computer interfaces: Definitions and principles},
editor = {Nick F. Ramsey and José del R. Millán},
series = {Handbook of Clinical Neurology},
publisher = {Elsevier},
volume = {168},
pages = {15-23},
year = {2020},
booktitle = {Brain-Computer Interfaces},
issn = {0072-9752},
doi = {https://doi.org/10.1016/B978-0-444-63934-9.00002-0},
url = {https://www.sciencedirect.com/science/article/pii/B9780444639349000020},
author = {Jonathan R. Wolpaw and José del R. Millán and Nick F. Ramsey},
}

@misc{li2020learning,
title={Learning Formation of Physically-Based Face Attributes},
  author={Li, Ruilong and Bladin, Karl and Zhao, Yajie and Chinara, Chinmay and Ingraham, Owen and Xiang, Pengda and Ren, Xinglei and Prasad, Pratusha and Kishore, Bipin and Xing, Jun and Li, Hao},
  booktitle={Proceedings of the IEEE/CVF conference on computer vision and pattern recognition},
  pages={3410--3419},
  year={2020}
}

@inproceedings{Li2020,
   title={An EEG-Based Multi-Modal Emotion Database with Both Posed and Authentic Facial Actions for Emotion Analysis},
   url={http://dx.doi.org/10.1109/FG47880.2020.00050},
   DOI={10.1109/fg47880.2020.00050},
   booktitle={2020 15th IEEE International Conference on Automatic Face and Gesture Recognition (FG 2020)},
   publisher={IEEE},
   author={Li, Xiaotian and Zhang, Xiang and Yang, Huiyuan and Duan, Wenna and Dai, Weiying and Yin, Lijun},
   year={2020},
   month=nov, pages={336–343} }

@article{Li2022, 
year = {2022}, 
title = {{A novel EEG decoding method for a facial-expression-based BCI system using the combined convolutional neural network and genetic algorithm}}, 
author = {Li, Rui and Liu, Di and Li, Zhijun and Liu, Jinli and Zhou, Jincao and Liu, Weiping and Liu, Bo and Fu, Weiping and Alhassan, Ahmad Bala}, 
journal = {Frontiers in Neuroscience}, 
doi = {10.3389/fnins.2022.988535}, 
pages = {988535}, 
volume = {16}
}

@inproceedings{qian2024gaussianavatarsphotorealisticheadavatars,
      title={GaussianAvatars: Photorealistic Head Avatars with Rigged 3D Gaussians}, 
     author={Qian, Shenhan and Kirschstein, Tobias and Schoneveld, Liam and Davoli, Davide and Giebenhain, Simon and Nie{\ss}ner, Matthias},
      booktitle={Proceedings of the IEEE/CVF Conference on Computer Vision and Pattern Recognition},
      pages={20299--20309},
      year={2024}

}

@book{coan2007handbook,
  title={Handbook of emotion elicitation and assessment},
  author={Coan, James A and Allen, John JB},
  year={2007},
  publisher={Oxford university press}
}

@article{MAHNOB,
  author={Soleymani, Mohammad and Lichtenauer, Jeroen and Pun, Thierry and Pantic, Maja},
  journal={IEEE Transactions on Affective Computing}, 
  title={A Multimodal Database for Affect Recognition and Implicit Tagging}, 
  year={2012},
  volume={3},
  number={1},
  pages={42-55},
  doi={10.1109/T-AFFC.2011.25}}

@inproceedings{rombach2021highresolution,
      title={High-Resolution Image Synthesis with Latent Diffusion Models}, 
        author={Rombach, Robin and Blattmann, Andreas and Lorenz, Dominik and Esser, Patrick and Ommer, Bj{\"o}rn},
      booktitle={Proceedings of the IEEE/CVF conference on computer vision and pattern recognition},
      pages={10684--10695},
      year={2022}

}

@article{emostim,
author = {Somarathna, Rukshani and Vuilleumier, Patrik and Mohammadi, Gelareh},
title = {EmoStim: A Database of Emotional Film Clips With Discrete and Componential Assessment},
year = {2024},
issue_date = {July-Sept. 2024},
publisher = {IEEE Computer Society Press},
address = {Washington, DC, USA},
volume = {15},
number = {3},
issn = {1949-3045},
url = {https://doi.org/10.1109/TAFFC.2023.3328900},
doi = {10.1109/TAFFC.2023.3328900},
journal = {IEEE Trans. Affect. Comput.},
month = jul,
pages = {1202–1212},
numpages = {11}
}

@article{li2023generalizable,
  title={Generalizable one-shot 3d neural head avatar},
  author={Li, Xueting and De Mello, Shalini and Liu, Sifei and Nagano, Koki and Iqbal, Umar and Kautz, Jan},
  journal={Advances in Neural Information Processing Systems},
  volume={36},
  pages={47239--47250},
  year={2023}
}

@inproceedings{tran2024voodoo,
  title={Voodoo 3d: Volumetric portrait disentanglement for one-shot 3d head reenactment},
  author={Tran, Phong and Zakharov, Egor and Ho, Long-Nhat and Tran, Anh Tuan and Hu, Liwen and Li, Hao},
  booktitle={Proceedings of the IEEE/CVF Conference on Computer Vision and Pattern Recognition},
  pages={10336--10348},
  year={2024}
}

@inproceedings{wang2021towards,
  title={Towards real-world blind face restoration with generative facial prior},
  author={Wang, Xintao and Li, Yu and Zhang, Honglun and Shan, Ying},
  booktitle={Proceedings of the IEEE/CVF conference on computer vision and pattern recognition},
  pages={9168--9178},
  year={2021}
}

@article {Willett2023.01.21.524489,
	author = {Willett, Francis R. and Kunz, Erin M. and Fan, Chaofei and Avansino, Donald T. and Wilson, Guy H. and Choi, Eun Young and Kamdar, Foram and Hochberg, Leigh R. and Druckmann, Shaul and Shenoy, Krishna V. and Henderson, Jaimie M.},
	title = {A high-performance speech neuroprosthesis},
	elocation-id = {2023.01.21.524489},
	year = {2023},
	doi = {10.1101/2023.01.21.524489},
	publisher = {Cold Spring Harbor Laboratory},
	URL = {https://www.biorxiv.org/content/early/2023/04/25/2023.01.21.524489},
	eprint = {https://www.biorxiv.org/content/early/2023/04/25/2023.01.21.524489.full.pdf},
	journal = {bioRxiv}
}


\end{document}